\documentclass[sigconf]{acmart}

\usepackage{enumitem}
\usepackage{balance} 
\usepackage{multirow}
\usepackage{graphicx}
\usepackage[dvipsnames, svgnames, x11names]{xcolor}
\usepackage{colortbl}

\usepackage{microtype}

\AtBeginDocument{%
  }

\copyrightyear{2026}
\acmYear{2026}
\setcopyright{cc}
\setcctype{by}
\acmConference[MM '26] {Proceedings of the 34th ACM International Conference on Multimedia}{November 10--14, 2026}{Rio de Janeiro, Brazil.}
\acmBooktitle{Proceedings of the 34th ACM International Conference on Multimedia (MM '26), November 10--14, 2026, Rio de Janeiro, Brazil}
\acmISBN{979-8-4007-2213-4/2026/11}
\acmDOI{10.1145/3767308.3835282}	

\begin{document}

\title{CoDS: Robust Collaborative Perception via Expert-driven Detection and BEV Segmentation}


\author{Jinlong Wang}
\affiliation{%
  \department{Guangdong Provincial Key Laboratory of Ultra High Definition Immersive Media Technology, Shenzhen Graduate School}
  \institution{Peking University}
  \city{Shenzhen}
  \country{China}}
\email{2601212983@stu.pku.edu.cn}

\author{Yuang Jia}
\affiliation{%
    \department{Guangdong Provincial Key Laboratory of Ultra High Definition Immersive Media Technology, Shenzhen Graduate School}
  \institution{Peking University}
  \city{Shenzhen}
  \country{China}}
\email{2601111556@stu.pku.edu.cn}

\author{Junhong Lin}
\affiliation{%
  \department{Guangdong Provincial Key Laboratory of Ultra High Definition Immersive Media Technology, Shenzhen Graduate School}
  \institution{Peking University}
  \city{Shenzhen}
  \country{China}}
\email{jhlin42in@gmail.com}

\author{Nannan Li}
\affiliation{%
 \department{School of Computer Science and Engineering}
 \institution{Macau University of Science and Technology}
 \city{Macau}
 \country{China}}
\email{nnli@must.edu.mo}

\author{Wei Gao}
\correspondingauthor
\authornote{Corresponding author: Wei Gao. This work was supported by National Science and Technology Major Project (2024ZD01NL00101), Natural Science Foundation of China (62271013), Guangdong Provincial Key Laboratory of Ultra High Definition Immersive Media Technology (2024B1212010006), Guangdong Province Pearl River Talent Program (2021QN020708), Guangdong Basic and Applied Basic Research Foundation (2024A1515010155), Shenzhen Science and Technology Program (JCYJ20240813160202004, JCYJ20230807120808017, SYSPG20241211173440004), Shenzhen Fundamental Research Program (GXWD20201231165807007-20200806163656003), and financially supported for Outstanding Talents Training Fund in Shenzhen.}
\affiliation{%
 \department{Guangdong Provincial Key Laboratory of Ultra High Definition Immersive Media Technology, Shenzhen Graduate School}
  \institution{Peking University}
  \city{Shenzhen}
  \country{China}}
\affiliation{%
  \institution{Peng Cheng Laboratory}
  \city{Shenzhen}
  \country{China}}
\email{gaowei262@pku.edu.cn}



\begin{abstract}
Collaborative perception breaks through single-view limitations via multi-agent information exchange. However, multi-source noise such as pose errors and communication delays degrades fusion feature quality, constraining perception performance. Joint training of detection and BEV segmentation provides a natural remedy, where segmented road regions help constrain target distributions and detection bounding boxes help recover ambiguous segmentation boundaries. To this end, we propose a robust Collaborative perception framework with expert-driven Detection and bev Segmentation (CoDS). To address spatial inconsistency in fusion quality, we first introduce the Collaborative Reliability Map (CoRM) to explicitly quantify feature quality distribution. Based on CoRM, we design the Semantic Mixture-of-Experts (S-MoE) module to extract differentiated features for inconsistent feature demands. Finally, to further mitigate feature noise degradation, the Bidirectional Task Complementary Interaction (BTCI) refines task-aware features through bidirectional injection. Extensive experiments on OPV2V and V2V4Real datasets show that our CoDS surpasses existing baselines on both tasks and maintains stable robustness under multi-source noise. Code: https://github.com/JinlongW128/CoDS and https://openi.pcl.ac.cn/OpenAIDriving/CoDS.
\end{abstract}










\maketitle

\section{Introduction}
A core prerequisite for L4/L5 autonomous driving is reliable, comprehensive environmental perception~\cite{1-huang2020autonomous,2-cui2024survey,3-hu2025collaborative,14-L4DR}. However, single-vehicle perception suffers from a limited field of view, facing inherent challenges in scenarios involving occlusions, distant targets, and viewpoint blind spots~\cite{4-han2023collaborative,5-liu2023towards,14-L4DR}. Collaboration breaks these viewpoint boundaries by multi-agent sharing, emerging as an important research direction in autonomous driving recently ~\cite{6-wang2025collaborative,7-xiang2023multi}. 


However, the collaborative fusion process faces interference from multi-source noise~\cite{15-catnet}. Sensor measurement errors and GPS/IMU positioning drift lead to deviations in inter-vehicle pose estimation~\cite{11-coalign}, while communication delays result in BEV features from different agents not originating from the same time instance~\cite{15-catnet,16-feaco}. These factors collectively cause spatial misalignment and noise accumulation during multi-agent BEV feature alignment, leading to local quality degradation of fusion features and constraining the performance ceiling of collaboration.

\begin{figure}[!t]
    \includegraphics[width=0.85\columnwidth]{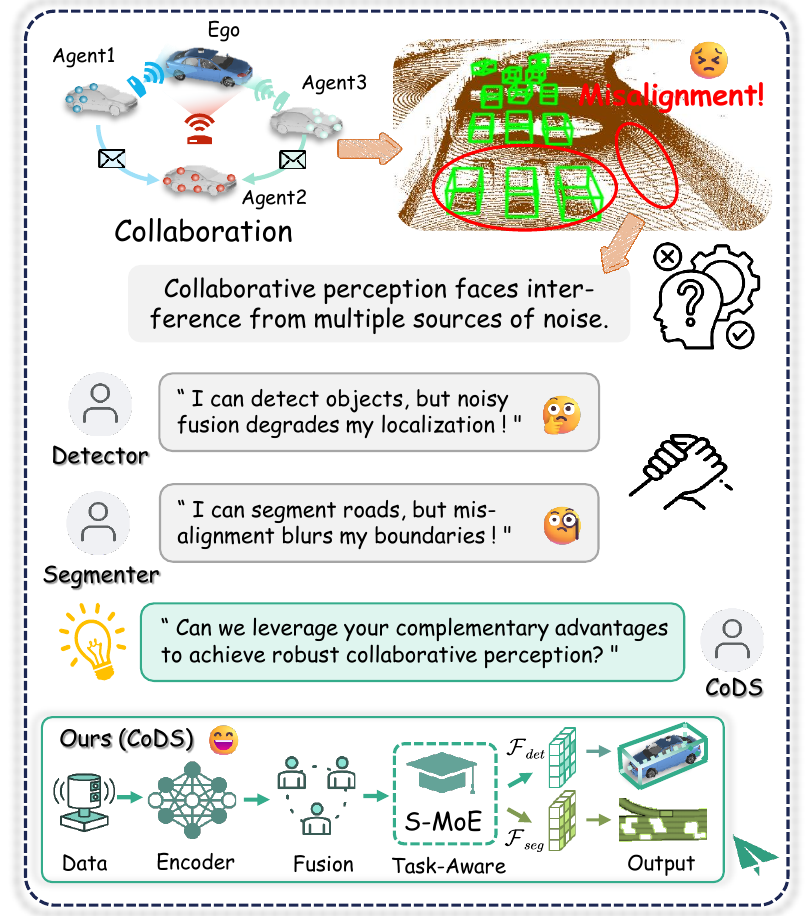}
  \caption{The motivation of our CoDS framework. 
  }
  \label{motivation}
  \Description{None.}
\end{figure}

We observe that stable road region responses from BEV semantic segmentation can provide spatial location priors for detection~\cite{17-pointbev}, constraining reasonable target distribution ranges; target bounding boxes from 3D object detection naturally delineate the spatial boundaries between object contours and road backgrounds~\cite{18-toshev2010object}, providing precise references for segmentation to recover confused boundaries. Based on these observations, a natural question arises, as shown in Fig.\ref{motivation}, \textit{can we leverage complementary supervision signals between tasks to achieve robust collaborative perception}?

However, most existing collaborative perception methods design independent pipelines to achieve 3D object detection~\cite{10-attfuse,11-coalign,8-v2xr,13-cosdh,15-catnet,48-c2e} and BEV semantic segmentation~\cite{12-cobevt,29-Communication-Efficient-seg,30-generative,31-toward}, resulting in significant redundancy in backbone computation. Although some works simultaneously handle both tasks~\cite{21-beverse,22-maskbev,23-maestro,19-cobevfusion,20-deepaccident,39-ICAV2X}, they either focus on single-agent scenarios without considering the unique challenges of collaborative settings~\cite{21-beverse,22-maskbev,23-maestro}, or perform both tasks in collaborative scenarios but lack cross-task bidirectional interaction~\cite{19-cobevfusion,20-deepaccident,39-ICAV2X,40-stamp}. 

It is worth noting that collaborative fusion features are a mixed representation of foreground and background information from multiple vehicles and viewpoints, exhibiting higher feature density and spatial heterogeneity~\cite{24-dsrc}. Therefore, achieving joint training for detection and segmentation faces two challenges. 
First, fusion quality demonstrates spatial inconsistency where well-aligned regions generate highly coherent and consistent multi-vehicle features. Conversely, regions with pose errors suffer from misalignment and noise that significantly degrade the resulting local feature quality.
The second issue is the inconsistent feature requirements between tasks~\cite{25-task,26-2014simultaneous}, as detection relies on locally accurate foreground boundary features, while segmentation relies on globally continuous background structural features.

To this end, we propose a robust \textbf{Co}llaborative perception framework with expert-driven \textbf{D}etection and BEV \textbf{S}egmentation (\textbf{CoDS}). The core idea is to use fusion reliability as a prior to drive task-specific feature extraction and cross-task interaction. 
To address the spatial inconsistency in fusion quality, we introduce the Collaborative Reliability Map (CoRM) to explicitly quantify the spatial quality distribution of BEV fusion features. Based on CoRM, we design the Semantic Mixture-of-Experts (S-MoE) module to extract task-aware features for inconsistent feature demands. Additionally, to further mitigate the noise-induced degradation of task-aware features, we design the Bidirectional Task Complementary Interaction (BTCI) to refine task-aware features via bidirectional injection.

We conduct comprehensive experiments on the OPV2V~\cite{10-attfuse} and V2V4Real~\cite{41-v2v4real} datasets. Results demonstrate that our CoDS surpasses existing collaborative perception baselines on both 3D object detection and BEV semantic segmentation tasks. Additionally, our CoDS maintains stable robustness under multi-source noise, validating the effectiveness of our expert-driven multi-task framework in combating fusion uncertainty. The main contributions of this paper are summarized as follows:

\begin{itemize}[leftmargin=10pt]
\item We propose the \textbf{CoDS}, a robust collaborative perception framework with expert-driven detection and BEV segmentation. It achieves robust perception by leveraging complementary interactions between detection and BEV segmentation.
\item We introduce the Collaborative Reliability Map (CoRM) to quantify the spatial inconsistency in fusion quality, and propose the Semantic Mixture-of-Experts (S-MoE) module for the inconsistent feature demands. To further mitigate the noise degradation, we design the Bidirectional Task Complementary Interaction (BTCI) to refine task-aware features.
\item Extensive experiments on OPV2V and V2V4Real datasets demonstrate that our CoDS surpasses existing baselines on both detection and bev segmentation tasks and maintains stable robustness under multi-source noise. 
\end{itemize}

\section{Related work}

\begin{figure*}[!t]
    \includegraphics[width=0.98\textwidth]{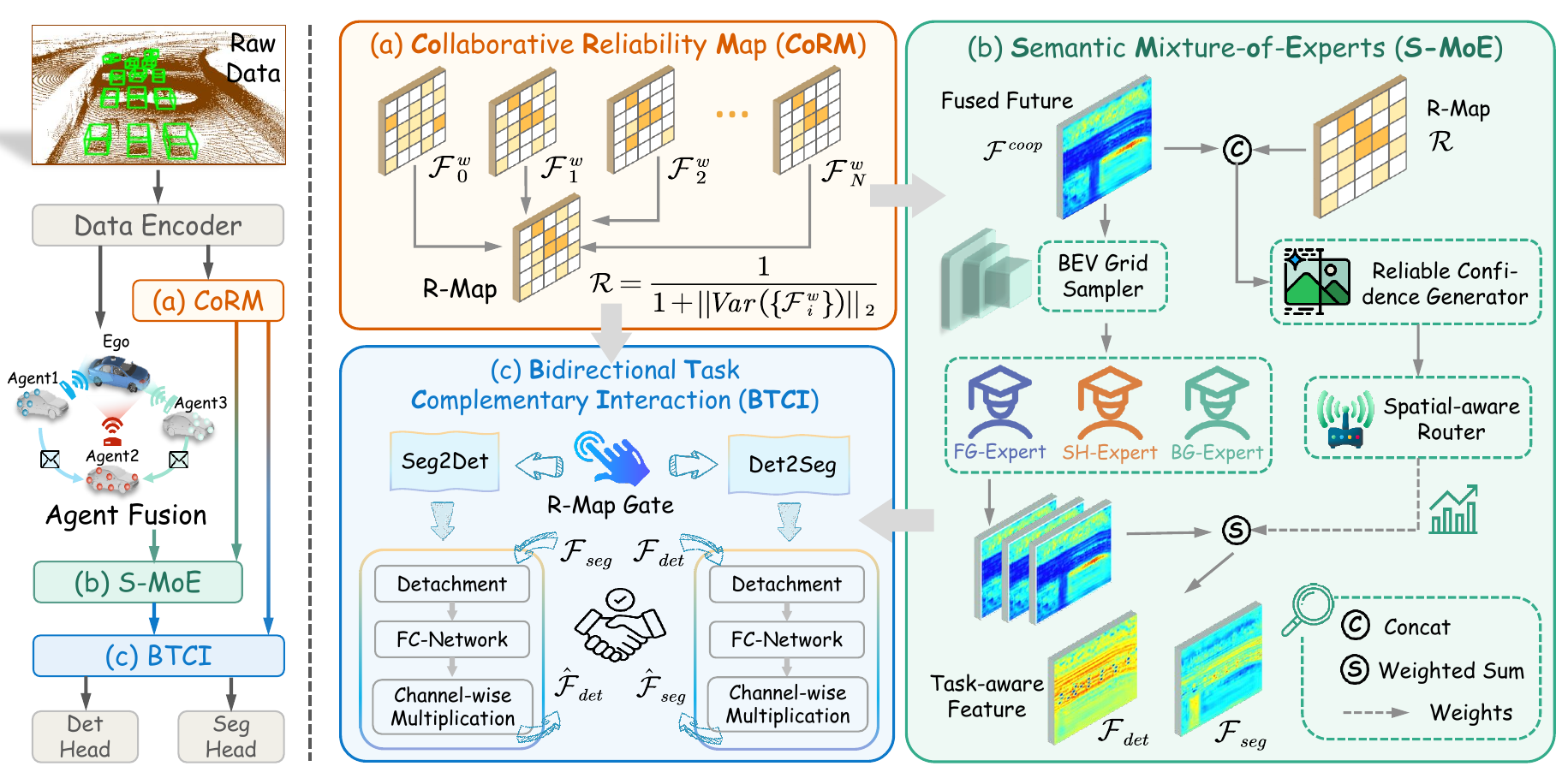}
  \centering
  \caption{\textbf{CoDS framework.} 
  Given $N$ CAV-encoded BEV features warped to ego coordinate via pose matrices, we get multi-agent features  $\{\mathcal{F}_i^{w} \in \mathbb{R}^{H \times W \times C}\}_{i=1}^{N}$. \textbf{CoRM} first analyzes pixel-wise consistency across these features, yielding a collaborative reliability map (R-Map) $\mathcal{R} \in (0,1]^{H \times W}$. The fused BEV feature $\mathcal{F}^{coop} \in \mathbb{R}^{H \times W \times C}$ is concatenated with $\mathcal{R}$ and passed to \textbf{S-MoE}. S-MoE uses $\mathcal{R}$ as a routing prior to distribute features among three semantic experts, producing $\mathcal{F}_{det}$ and $\mathcal{F}_{seg}$. Finally, \textbf{BTCI} uses $\mathcal{R}$ as a gating signal for bidirectional interaction between $\mathcal{F}_{det}$ and $\mathcal{F}_{seg}$, yielding refined $\hat{\mathcal{F}}_{det}$ and $\hat{\mathcal{F}}_{seg}$ for detection and segmentation heads.
  }
  \label{framework}
  \Description{None.}
\end{figure*}

\subsection{Collaborative Perception}
Collaborative perception overcomes the fixed FOV limit of single-vehicle perception by sharing complementary information  among agents. Existing fusion paradigms can be categorized into early~\cite{disconet,cooper,Double-M_Quantification}, intermediate~\cite{10-attfuse,
11-coalign,8-v2xr,12-cobevt,13-cosdh,15-catnet,28-cobevmoe,32-bm2cp,33-coca3d,34-eimc}, and late fusion~\cite{mash,mamp} based on the fusion stage. Intermediate fusion offers the best trade-off between performance and cost, so most works study efficient feature fusion~\cite{9-where2comm}, communication strategies~\cite{13-cosdh}, and pose alignment~\cite{11-coalign}. However, intermediate fusion relies on feature-level alignment. This makes it sensitive to multi-source noise. Sensor errors, GPS/IMU drift, and latency cause spatial misalignment and noise accumulation in BEV alignment~\cite{11-coalign,16-feaco,27-ermvp,15-catnet}, degrading local fusion quality and limiting collaboration performance. Existing robust methods mainly handle pose correction~\cite{11-coalign,16-feaco}, compression and denoising~\cite{27-ermvp}, and synchronization~\cite{15-catnet}. However, these methods neglect the spatial inconsistency of fusion quality, where well-aligned regions coexist with error-prone ones, and lack explicit mechanisms to model this quality distribution and handle it adaptively, limiting robustness under real-world uncertainties.

\subsection{Multi-Task Learning for Perception}
Existing collaborative works have made much progress in 3D detection~\cite{10-attfuse,11-coalign,8-v2xr,13-cosdh,15-catnet,32-bm2cp,33-coca3d,34-eimc} and BEV segmentation~\cite{12-cobevt,29-Communication-Efficient-seg,30-generative, 31-toward} separately. However, they are often treated as independent pipelines, causing redundant backbone computation and missing their complementarity. Some works jointly handle both tasks in single-vehicle settings~\cite{35-m2bev,36-m3net,21-beverse,23-maestro, 22-maskbev,37-diffusion,38-bevfusion}, but they ignore challenges unique to collaborative fusion. In fusion, BEV features aggregate info from multiple vehicles and viewpoints, showing higher spatial variety than single-agent features~\cite{24-dsrc}. This leads to uneven fusion quality and different task needs. A few works address both tasks in collaborative settings~\cite{19-cobevfusion,20-deepaccident,39-ICAV2X,40-stamp,28-cobevmoe}. But they either process each task alone without interaction~\cite{19-cobevfusion,20-deepaccident,28-cobevmoe,40-stamp}, or use segmentation only as a communication filter, not as a complement to detection~\cite{39-ICAV2X}. To this end, we aim to fully exploit their bidirectional complementarity for robust multi-task collaboration.
\section{Methodology}

\subsection{Preliminaries}
In collaborative perception, ego vehicle and surrounding Connected Automated Vehicles (CAVs) share intermediate BEV features for perception. Each agent $i$ extracts the local BEV feature $\mathcal{F}_i \in \mathbb{R}^{H \times W \times C}$, which are spatially aligned by pose transformation and fused into a collaborative BEV feature $\mathcal{F}^{coop} \in \mathbb{R}^{H \times W \times C}$. However, pose estimation errors from sensor noise and GPS/IMU drift, along with temporal misalignment from communication latency, introduce spatial inconsistencies during feature alignment, leading to heterogeneous fusion quality across the BEV space.

To achieve robust collaborative perception, we jointly optimize 3D object detection and BEV semantic segmentation. For 3D object detection, our model regresses a set of 3D bounding boxes $\mathcal{B}_{det} = \{b_i\}_{i=1}^{N_b}$, where each $b_i \in \mathbb{R}^7$ encodes position, dimensions, and orientation. For BEV semantic segmentation, it predicts per-pixel semantic labels $\mathcal{S}_{seg} \in \mathbb{R}^{H \times W \times N_c}$ over $N_c$ categories (e.g., road, lane, and dynamic objects).

\subsection{Overall Design}
As shown in Fig.~\ref{framework}, we propose \textbf{CoDS}, an expert-driven framework for robust collaborative perception with detection and BEV segmentation. First, to tackle the spatial inconsistency of fusion quality, the \textbf{Co}llaborative \textbf{R}eliability \textbf{M}ap (CoRM) generation  (Fig.~\ref{framework} (a)) explicitly quantifies the spatial quality distribution of collaborative BEV features. Based on CoRM, the \textbf{S}emantic \textbf{M}ixture-\textbf{o}f-\textbf{E}xperts (S-MoE) module (Fig.~\ref{framework} (b)) performs task-aware feature extraction for inconsistent feature demands. Moreover, to further mitigate the noise-induced degradation, the \textbf{B}idirectional \textbf{T}ask \textbf{C}omplementary \textbf{I}nteraction (BTCI) module (Fig.~\ref{framework} (c)) refines features via bidirectional injection. More detailed implementation is provided below.

\subsection{Collaborative Reliability Map}
Collaborative perception aligns agent BEV features via pose transform before fusion. Pose estimation errors and communication delays introduce spatial misalignment during warping, leading to heterogeneous fusion quality across the BEV space: well-aligned regions exhibit high inter-agent feature consistency and reliable collaborative information, whereas misaligned regions suffer from spatial offset and noise accumulation. However, existing methods treat fused features uniformly, ignoring spatial quality modeling.

To address this, we propose the \textbf{Co}llaborative \textbf{R}eliability \textbf{M}ap (\textbf{CoRM}) generation module, which explicitly quantifies the spatial quality distribution of collaborative BEV features by measuring pixel-wise inter-agent feature consistency. Regions of high consistency indicate successful alignment and trustworthy collaborative information, while low-consistency regions signal potential misalignment and warrant reliability-aware processing.

\textbf{Spatial Consistency Measurement.}
Given $N$ surrounding agents participating in collaborative perception, each agent $i$ extracts its local BEV feature $\mathcal{F}_i \in \mathbb{R}^{H \times W \times C}$ and transforms it to the ego vehicle's coordinate system via spatial warping, yielding warped features $\{\mathcal{F}_1^w, \mathcal{F}_2^w, \ldots, \mathcal{F}_N^w\}$. As shown in Fig.~\ref{framework} (a), we stack these features along the agent dimension and compute the pixel-wise variance across agents to capture spatial inconsistency. Specifically, at each pixel-wise spatial location $(h, w)$, we obtain a variance vector across all channels:
\begin{equation}
\begin{aligned}
\mathcal{V}(h, w) = \text{Var}(\{\mathcal{F}_i^w(:, h, w)\}_{i=1}^{N}) \in \mathbb{R}^{C},
\end{aligned}
\end{equation}
where $\mathcal{V}(h, w)$ represents the channel-wise variance vector at location $(h, w)$. Higher variance indicates greater feature disagreement among agents, signaling potential spatial misalignment.
To aggregate the channel-wise variance into a scalar spatial inconsistency measure, we compute the L2 norm:
\begin{equation}
\begin{aligned}
\varphi(h, w) = \|\mathcal{V}(h, w)\|_2 \in \mathbb{R}.
\end{aligned}
\end{equation}

\begin{figure}[!t]
\centering
\includegraphics[width=0.9\linewidth]{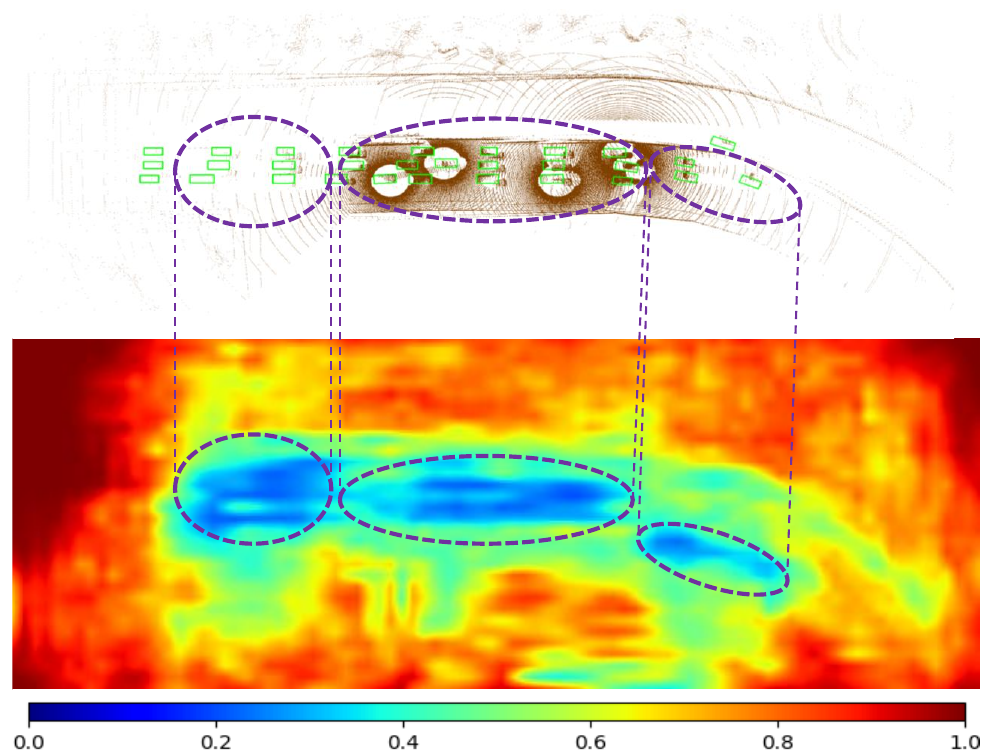}
\caption{Visualization of R-Map under Gaussian pose noise $\eta$\,$(2.0\,\text{m},\,2.0^\circ)$ on OPV2V. Low-reliability regions (\textcolor{Blue}{blue}) coincide with areas of dense vehicle presence, where pose noise induces severe feature misalignment.}
\Description{None.}
\label{corm}
\end{figure}

\textbf{R-Map Generation.}
To convert spatial inconsistency into reliability, we apply an inverse transformation. The collaborative reliability map (R-Map) is generated as:
\begin{equation}
\begin{aligned}
\mathcal{R}(h, w) = \frac{1}{1 + \varphi(h, w)} \in (0, 1],
\end{aligned}
\end{equation}
where larger variance $\varphi(h, w)$ yields lower reliability scores, and vise versa. The resulting reliability map $\mathcal{R} \in \mathbb{R}^{H \times W \times 1}$ provides pixel-wise confidence estimation of the collaborative feature $\mathcal{F}^{coop}$.

As illustrated in Fig.~\ref{corm}, regions with high reliability scores correspond to well-aligned collaborative features with consistent multi-agent information, while low-reliability regions indicate spatial misalignment and noise-corrupted features. This explicit spatial quality quantification serves as a prior to guide subsequent task-specific feature extraction in the S-MoE.

\begin{figure}[!t]
\centering
\includegraphics[width=1.0\linewidth]{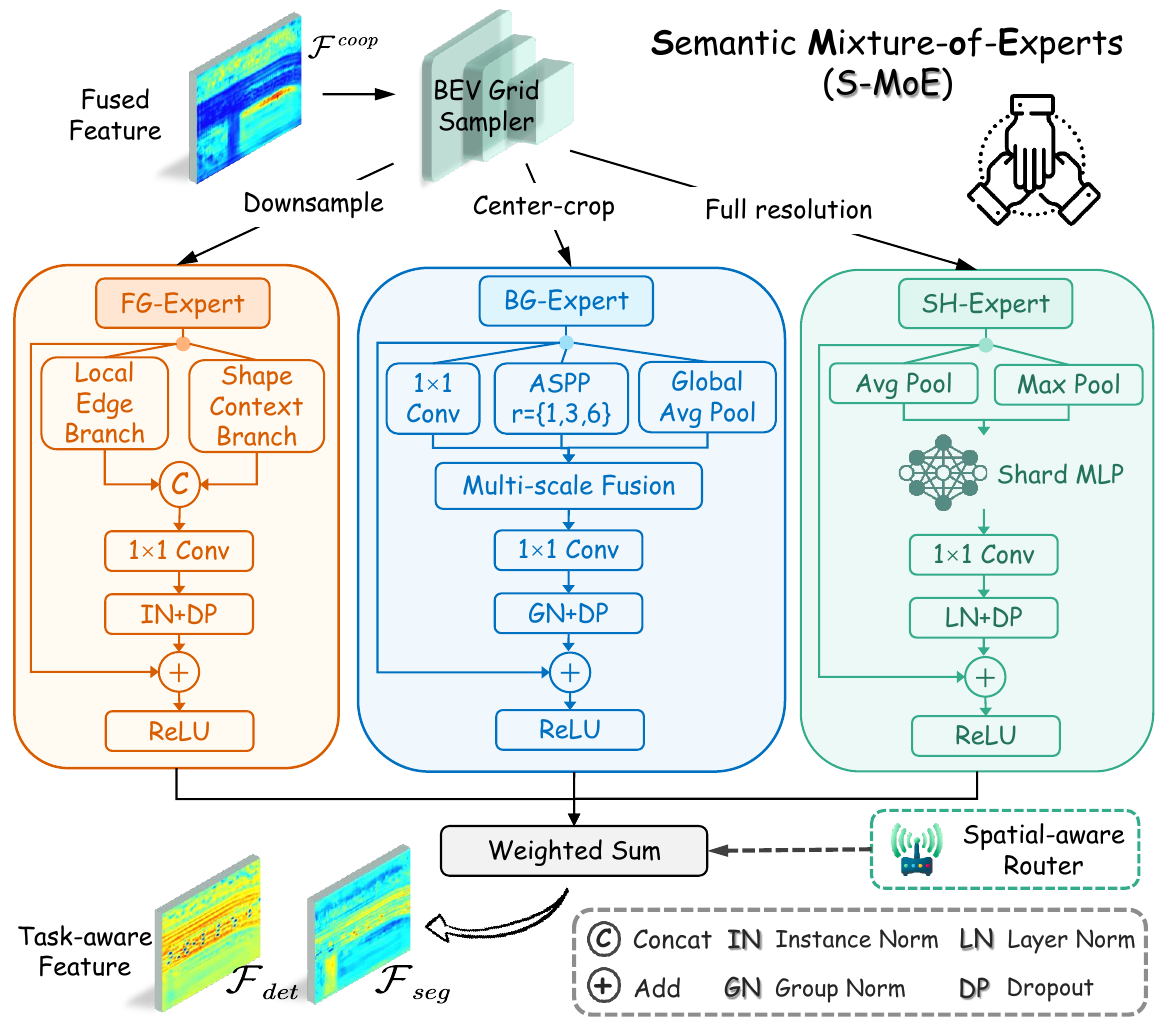}
\caption{Semantic Mixture-of-Experts (S-MoE). The fused feature $\mathcal{F}^{coop}$, concatenated with the reliability map $\mathcal{R}$, is routed to the semantic experts via a reliability-guided gating mechanism, producing task-aware features $\mathcal{F}_{det}$ and $\mathcal{F}_{seg}$.}
\label{fig:expert}
\Description{None.}
\end{figure}

\subsection{Semantic Mixture-of-Experts}

The fused collaborative feature $\mathcal{F}^{coop}$ aggregates multi-view foreground and background, exhibiting two forms of spatial heterogeneity that a single extractor cannot adequately handle. First, detection and segmentation have different needs: detection needs fine boundaries in sparse foreground, while BEV segmentation needs spatial continuity for road topology. Second, pose errors cause fusion quality to vary spatially. Routing low-reliability regions blindly to experts may amplify noise. Thus, we propose the \textbf{S}emantic \textbf{M}ixture-\textbf{o}f-\textbf{E}xperts (\textbf{S-MoE}), which uses reliability-guided, task-aware extraction to address both issues.

Since dynamic object segmentation shares semantic targets with detection, we design experts by semantics, not tasks. As shown in Fig.~\ref{fig:expert}, the foreground expert (\textbf{FG-Expert}) extracts fine boundaries via two parallel branches: a $3{\times}3$ convolution for local edges and a $5{\times}5$ depthwise separable convolution that encodes shape context. Their outputs are summed and passed through a residual path with InstanceNorm for sparse activations. To cover both lane markings and intersections, the background expert (\textbf{BG-Expert})  adopts Atrous Spatial Pyramid Pooling (ASPP)~\cite{46-ASPP} with rates $\{1, 3, 6\}$ to aggregate multi-scale road patterns, with GroupNorm for stable dense training. The shared expert (\textbf{SH-Expert}) performs global channel recalibration via dual-path SE attention~\cite{47-SENet}, avoiding spatial bias and serving as a fallback for uncertain regions. All expert outputs are bilinearly re-aligned to $H \times W$ with residual correction, yielding $\mathcal{E}_{fg}$, $\mathcal{E}_{bg}$, $\mathcal{E}_{sh} \in \mathbb{R}^{C \times H \times W}$.

Given distinct foreground and background distributions, we adopt multi-scale BEV grid sampling to give each expert tailored inputs. Specifically, the FG-Expert gets a downsampled full-range feature to broaden context for sparse objects. BG-Expert uses a cropped and upsampled central region to preserve fine details like lane markings. SH-Expert directly uses full-resolution $\mathcal{F}^{coop}$.

We then design a spatial-aware router for the three experts at each spatial location. As shown in Fig.~\ref{framework} (b), to allow the router to identify not only what semantic content a region contains but also how reliable it is, we first design a reliable confidence generator $\mathcal{H}_{conf}$ that introduces the collaborative reliability map $\mathcal{R}$ as an additional prior into the confidence estimation process. The generator concatenates $\mathcal{F}^{coop}$ with $\mathcal{R}$ and feeds the result into two parallel lightweight convolutional branches, producing foreground confidence $\mathcal{P}_{fg}$ and background confidence $\mathcal{P}_{bg}$:
\begin{equation}
  \mathcal{P}_{fg},\, \mathcal{P}_{bg}
  = \mathcal{H}_{conf}\!\left([\mathcal{F}^{coop},\, \mathcal{R}]\right).
\end{equation}

The  spatial-aware router combines a semantic routing prior $\{\boldsymbol{\pi}_{det},\boldsymbol{\pi}_{seg}\} $ with a learned routing signal $\{\boldsymbol{\lambda}_{det}, \boldsymbol{\lambda}_{seg}\} $ . For detection, the FG-Expert weight is proportional to $\mathcal{P}_{fg}$, with the SH-Expert compensating in low-reliability regions. For segmentation, all three experts participate in routing according to foreground confidence, background confidence, and positional uncertainty:
\begin{equation}
  \begin{aligned}
    \boldsymbol{\pi}_{det}(i,j)
    &= \bigl[\mathcal{P}_{fg},\; 0,\; \mathcal{P}_{bg}\bigr], \\
    \boldsymbol{\pi}_{seg}(i,j)
    &= \bigl[\mathcal{P}_{fg},\; \mathcal{P}_{bg},\;
      1 - \max(\mathcal{P}_{fg}, \mathcal{P}_{bg})\bigr],
  \end{aligned}
\end{equation}
where $\boldsymbol{\pi}_\tau(i,j) \in \mathbb{R}^3$ denotes the semantic routing prior at spatial location $(i,j)$ for task $\tau$, with entries ordered as $\{{fg},\, {bg},\, {sh}\}$ and $\mathcal{P}_{bg} = 1 - \mathcal{P}_{fg}$ represents the background confidence derived from CoRM.
And the learned routing signal $\boldsymbol{\lambda}_\tau \in \mathbb{R}^{H \times W \times 3}$ is obtained from a $3{\times}3$ local spatial attention branch conditioned on the task-modulated feature.  The two signals are combined via a task-specific learnable scalar $\alpha_\tau$ before Softmax normalization:
\begin{equation}
  \mathcal{G}_\tau
  = \mathrm{Softmax}\!\bigl(
      \alpha_\tau \cdot \boldsymbol{\pi}_\tau
      + (1 - \alpha_\tau) \cdot \boldsymbol{\lambda}_\tau
    \bigr),
  \quad \tau \in \{\mathrm{det},\, \mathrm{seg}\},
\end{equation}
where $\mathcal{G}_\tau \in \mathbb{R}^{H \times W \times 3}$ is the normalized gating weight map for task $\tau$, and $\alpha_\tau$ is initialized to $0.7$ for detection and $0.45$ for segmentation. This is to emphasize the semantic prior in the detection task, while the segmentation relies more on the learned signal.

Finally, the task-specific features are obtained by spatially weighted summation
of the three expert outputs:
\begin{equation}
  \mathcal{F}_\tau
  = \sum_{k \in \{fg,\, bg,\, sh\}}
    \mathcal{G}_\tau^{(k)} \odot \mathcal{E}_k,
  \quad \tau \in \{\mathrm{det},\, \mathrm{seg}\},
\end{equation}
where $\odot$ denotes element-wise multiplication.
The aggregated features are then refined by a lightweight task adaptor
($1{\times}1$ residual convolution) to produce task-aware $\mathcal{F}_{det}$ and
$\mathcal{F}_{seg}$.

\subsection{Bidirectional Task Complementary Interaction}

Detection features carry precise boundary knowledge around sparse objects, while segmentation features contain stable prior knowledge of large-scale road structures, providing a natural opportunity for bidirectional complementarity. Although S-MoE can generate task-specific features $\mathcal{F}_{det}$ and $\mathcal{F}_{seg}$, these two features are still susceptible to noise interference in low-reliability areas, and simple cross-task interaction may pose a risk of training instability.
To this end, we propose the \textbf{B}idirectional \textbf{T}ask \textbf{C}omplementary
\textbf{I}nteraction \textbf{(BTCI)} module to refine the task-aware features through the bidirectional feature injection.

As shown in Fig.~\ref{framework} (c), BTCI extracts channel weights from the source via global average pooling and lightweight full-connection network and then performs reliability-gated channel-wise multiplication on target. It also detaches gradients bidirectionally  to ensure independent branch convergence. 

Taking the \textit{det2seg} direction as an example, a channel descriptor is extracted from $\mathcal{F}_{det}$ by global average pooling and passed through a two-layer Fully Connected (FC) Network to produce channel weights:
\begin{equation}
    \mathbf{W} = \sigma\!\left(\mathbf{W}_2\,\delta\!\left(\mathbf{W}_1\,
    \frac{1}{HW}\sum_{h,w}\mathcal{F}_{det}^{h,w}\right)\right) \in (0,1)^C,
    \label{eq:channel_weight}
\end{equation}
where $C$ is channel, $\mathbf{W}_1 \in \mathbb{R}^{(C/\rho) \times C}$ and $\mathbf{W}_2 \in \mathbb{R}^{C \times (C/\rho)}$ denote the FC weight matrices with channel reduction ratio $\rho=8$, $\sigma(\cdot)$ is the sigmoid function, and $\delta(\cdot)$ is the ReLU activation. 
The weights and bias of $\mathbf{W}_2$ are zero-initialized, ensuring $\mathbf{W} \equiv \mu$ at the start of training, where $\mu = 0.5$ is the sigmoid midpoint. This ensures that in the early stages of training, BTCI starts learning from the identity mapping.

Then, BTCI computes a scalar gate $\beta = \frac{1}{HW}\sum_{h,w}\mathcal{R}_{h,w} \in (0,1]$ from the reliability map $\mathcal{R}$ to uniformly control the interaction strength in both directions. Subsequently, the channel weights modulate the target feature via multiplicative residual injection:
\begin{equation}
    \hat{\mathcal{F}}_{seg} = \mathcal{F}_{seg} \odot
    \left(1 + \beta \cdot \left(\mathbf{W} - \mu\right)\right),
    \label{eq:hint_injection}
\end{equation}
where $\odot$ denotes element-wise multiplication. When $\beta \to 0$, the injection term vanishes and the module reduces to an identity mapping in low-reliability regions. BTCI applies the same mechanism symmetrically in the \textit{seg2det} direction, sharing $\beta$ across both directions while maintaining separate FC parameters. 

Stop-gradient is applied to the source feature in each direction to prevent cross-task gradient interference, yielding the complete bidirectional formulation:
\begin{align}
    \hat{\mathcal{F}}_{seg} &= \mathcal{F}_{seg} \odot
    \left(1 +  \cdot \beta \cdot
    \left(\mathbf{W}_{d2s}\!\left(\mathrm{sg}[\mathcal{F}_{det}]\right)
    - \mu\right)\right), \label{eq:btci_seg}\\
    \hat{\mathcal{F}}_{det} &= \mathcal{F}_{det} \odot
    \left(1 +  \cdot \beta \cdot
    \left(\mathbf{W}_{s2d}\!\left(\mathrm{sg}[\mathcal{F}_{seg}]\right)
    - \mu\right)\right), \label{eq:btci_det}
\end{align}
where $\mathbf{W}_{d2s}$ and $\mathbf{W}_{s2d}$ denote the FC networks for the \textit{det2seg} and \textit{seg2det} directions respectively, and $\mathrm{sg}[\cdot]$ denotes the stop-gradient operation. The final refined features $\hat{\mathcal{F}}_{det}$ and $\hat{\mathcal{F}}_{seg}$ are input into task heads to complete detection and BEV segmentation. 

\subsection{Loss Function}
The total training loss $\mathcal{L}$ is defined as:
\begin{equation}
\mathcal{L} = \zeta_{{cls}}\,\mathcal{L}_{{cls}} + \zeta_{{reg}}\,\mathcal{L}_{{reg}} + \zeta_{{seg}}\,\mathcal{L}_{{seg}},
\end{equation}
where $\mathcal{L}_{{cls}}$ and $\mathcal{L}_{{reg}}$ are the focal loss for classification and smooth-$\ell_1$ loss for box regression with $\zeta_{{cls}}{=}1.0$, $\zeta_{{reg}}{=}2.0$. And $\mathcal{L}_{\text{seg}} = \zeta_{s}\,\mathcal{L}_{{s}} + \zeta_{d}\,\mathcal{L}_{{d}}$ supervises static and dynamic semantics with $\zeta_{s}{=}1.0$, $\zeta_{d}{=}2.0$.

\section{Experiments}

\subsection{Experimental Settings}
\textbf{Datasets.}
We validate our CoDS on two datasets: OPV2V~\cite{10-attfuse} and V2V4Real~\cite{41-v2v4real}. OPV2V is a large-scale open-source simulated dataset, which provides annotations for both 3D object detection and BEV semantic segmentation. V2V4Real is a large-scale real-world dataset with detection annotations, enabling evaluation of generalization in real-world environments.

\textbf{Baselines Selection.}
We evaluate CoDS against two groups of baselines. The first includes representative single-task cooperative perception methods for either detection or BEV segmentation~\cite{10-attfuse,9-where2comm,11-coalign,13-cosdh,12-cobevt}. The second covers multi-task frameworks spanning single-agent~\cite{21-beverse} and cooperative settings~\cite{19-cobevfusion,39-ICAV2X}.

\textbf{Implementation Details.} 
Our CoDS is built upon the intermediate fusion strategy of AttFuse~\cite{10-attfuse} and trained and evaluated on both OPV2V~\cite{10-attfuse} and V2V4Real~\cite{41-v2v4real}. The LiDAR range for detection is set to $x \in [-140.8, 140.8]$ m and $y \in [-38.4, 38.4]$ m. For BEV semantic segmentation, we follow the annotation range provided by OPV2V, configured as $x, y \in [-51.2, 51.2]$ m. V2V4Real lacks BEV segmentation annotations, so we use detection GT boxes to generate dynamic ground truth, validating segmentation effectiveness under real-world noise.
The learning rate is reduced by a factor of 0.1 at epochs 10 and 20, with training running for 30 epochs in total. The batch size is set to 2, and the maximum number of CAVs per frame is capped at 5. To ensure fair comparison, all baselines are retrained under the identical experimental configuration as ~\cite{10-attfuse}, and all results are obtained from two NVIDIA RTX 4090 GPUs.
More details are provided in the \textit{{Appx.}}. 

\begin{table}[!t]
\centering
\caption{Results on the OPV2V. Best results are in \textbf{bold}.}
\label{tab:opv2v}
\resizebox{\columnwidth}{!}{%
\begin{tabular}{c|ccc | ccc}
\hline
\multirow{2}{*}{\textbf{Method}}  & \multicolumn{3}{c|}{\textbf{Detection}} & \multicolumn{3}{c}{\textbf{Segmentation}} \\
\cline{2-7}
 & \textbf{AP$@$0.3} & \textbf{AP$@$0.5} & \textbf{AP$@$0.7} & \textbf{Dynamic} & \textbf{Road} & \textbf{Lane} \\
\hline
AttFuse~\cite{10-attfuse}  & 89.94 & 89.41 & 83.65  & 71.20 & 63.97 & 48.08 \\
Where2comm~\cite{9-where2comm} & 90.19 & 89.20 & 80.55 &  70.19 & 63.43 & 47.36 \\
CoBEVT~\cite{12-cobevt} & 92.09 & 91.09 & 80.95& 73.97 & 65.53 & 50.84 \\
CoAlign~\cite{11-coalign}  & 89.27 & 88.85 & 82.73  & 71.57 & 63.88 & 48.33 \\
CoSDH~\cite{13-cosdh}  & 91.06 & 90.54 & 82.48 & 72.54 & 62.35 & 46.31 \\
\hline
Ours (CoDS)  & \textbf{94.64} & \textbf{94.05} & \textbf{86.88} &  \textbf{76.47} & \textbf{69.38} & \textbf{55.19} \\
\hline
\end{tabular}
}
\end{table}

\begin{table}[!t]
\centering
\caption{Results on the V2V4Real. Best results are in \textbf{bold}.}
\label{tab:v2v4real}
\resizebox{1.0\columnwidth}{!}{%
\begin{tabular}{c |c| ccc |c}
\hline
\textbf{Method} & \textbf{Public} & \textbf{AP$@$0.3} & \textbf{AP$@$0.5} & \textbf{AP$@$0.7} &  \textbf{Dynamic}\\
\hline
AttFuse~\cite{10-attfuse} & ICRA 2022 & 66.45 & 59.23 & 31.43 & 54.29 \\
Where2comm~\cite{9-where2comm} & NeurIPS 2022& 65.11 & 58.77 & 31.21 & 63.44\\
CoBEVT~\cite{12-cobevt} & CoRL 2022 & 68.05 & 58.78 & 31.70 & 64.94\\
CoAlign~\cite{11-coalign} & ICRA 2023  & 69.45 & 62.02 & 31.81 & 57.58\\
CoSDH~\cite{13-cosdh} & CVPR 2025   & 67.64 & 61.42 & 31.00 & 61.58\\
\hline
Ours (CoDS) & - & \textbf{72.63} & \textbf{66.04} & \textbf{38.33} & \textbf{65.32} \\
\hline
\end{tabular}
}
\end{table}

\subsection{Performance Results}

\textbf{Results on the OPV2V.}
Table~\ref{tab:opv2v} reports the 3D detection and BEV segmentation results on OPV2V. Overall, CoDS achieves the best performance across all metrics. For detection, it attains AP$@$0.3, AP$@$0.5, and AP$@$0.7 of 94.64, 94.05, and 86.88, surpassing AttFuse~\cite{10-attfuse} by 4.70, 4.64, and 3.23 points. Notably, CoDS achieves the most consistent improvement at AP$@$0.7, indicating that segmentation knowledge helps refine bounding box localization. For segmentation, CoDS outperforms AttFuse by 5.27, 5.41, and 7.11 points on Dynamic, Road, and Lane IoU, respectively. The notable gain on Dynamic shows that detection provides useful cues for segmenting moving objects, confirming the complementary benefits of joint multi-task modeling.

\textbf{Results on the V2V4Real.}
Table~\ref{tab:v2v4real} presents results on the real-world V2V4Real dataset. CoDS consistently outperforms all baselines across detection metrics, achieving 72.63, 66.04, and 38.33 for AP$@$0.3, AP$@$0.5, and AP$@$0.7, respectively. The pronounced improvement at the strict AP$@$0.7 metric verifies CoDS's robustness under real-world sensor noise and pose perturbations. For segmentation, CoDS also achieves the best Dynamic IoU of 65.32, further validating the effectiveness of cross-task complementarity in challenging real-world scenarios.

\begin{table}[!t]
\centering
\caption{Ablation study of each proposed module on OPV2V. 
``Baseline (only-det)'' and ``Baseline (only-seg)'' denote single-task detection and segmentation models. 
Best in \textbf{bold}.}
\label{tab:ablation}
\resizebox{\columnwidth}{!}{%
\begin{tabular}{c | c | ccc}
\hline
\textbf{Module} & \textbf{Test (AP$@$0.3/0.5/0.7)} & \textbf{Dynamic} & \textbf{Road} & \textbf{Lane} \\
\hline
Baseline~\cite{10-attfuse} (only-det) & 89.94 / 89.41 / 83.65 & -     & -     & -     \\
Baseline~\cite{10-attfuse} (only-seg) & -                      & 71.20 & 63.97 & 48.08 \\
\hline
+Seg Head           & 92.98 / 92.01 / 84.59  & 71.05 & 62.39 & 46.82 \\
+S-MoE              & 93.32 / 92.62 / 86.03  & 73.73 & 65.13 & 50.44 \\
+R-Map              & 94.22 / 93.56 / 85.68  & 75.82 & 67.44 & 53.11 \\
+BTCI (Ours)        & \textbf{94.64 / 94.05 / 86.88}  & \textbf{76.47} & \textbf{69.38} & \textbf{55.19} \\
\hline
\end{tabular}
}
\end{table}

\subsection{Ablation Study}
We conduct an incremental ablation study (Table~\ref{tab:ablation}) on OPV2V. Single-task AttFuse~\cite{10-attfuse} serves as the baseline. First, adding a parallel segmentation head to detection (+Seg Head) improves detection at all IoU thresholds, confirming joint training helps detection. But segmentation slightly degrades relative to the only-seg, showing a plain head cannot handle the different feature needs of both tasks. Second, adding S-MoE (+S-MoE) gains both tasks, with AP$@$0.7 rises to 86.03 and all segmentation IoUs improve. This shows task-specific extraction effectively alleviates the multi-task bottleneck. Third, adding R-Map (+R-Map) provides spatial quality priors to guide routing, steadily improving both tasks and proving the need to model fusion reliability. Finally, adding BTCI (+BTCI) achieves the best results, raising AP$@$0.7 by 2.29 and Dynamic IoU by 5.42 over the naive multi-task baseline (+Seg Head). This confirms the complementary gains from bidirectional feature injection.

\subsection{Analysis Experiments}

\begin{figure*}[!t]
    \centering
    \includegraphics[width=\textwidth]{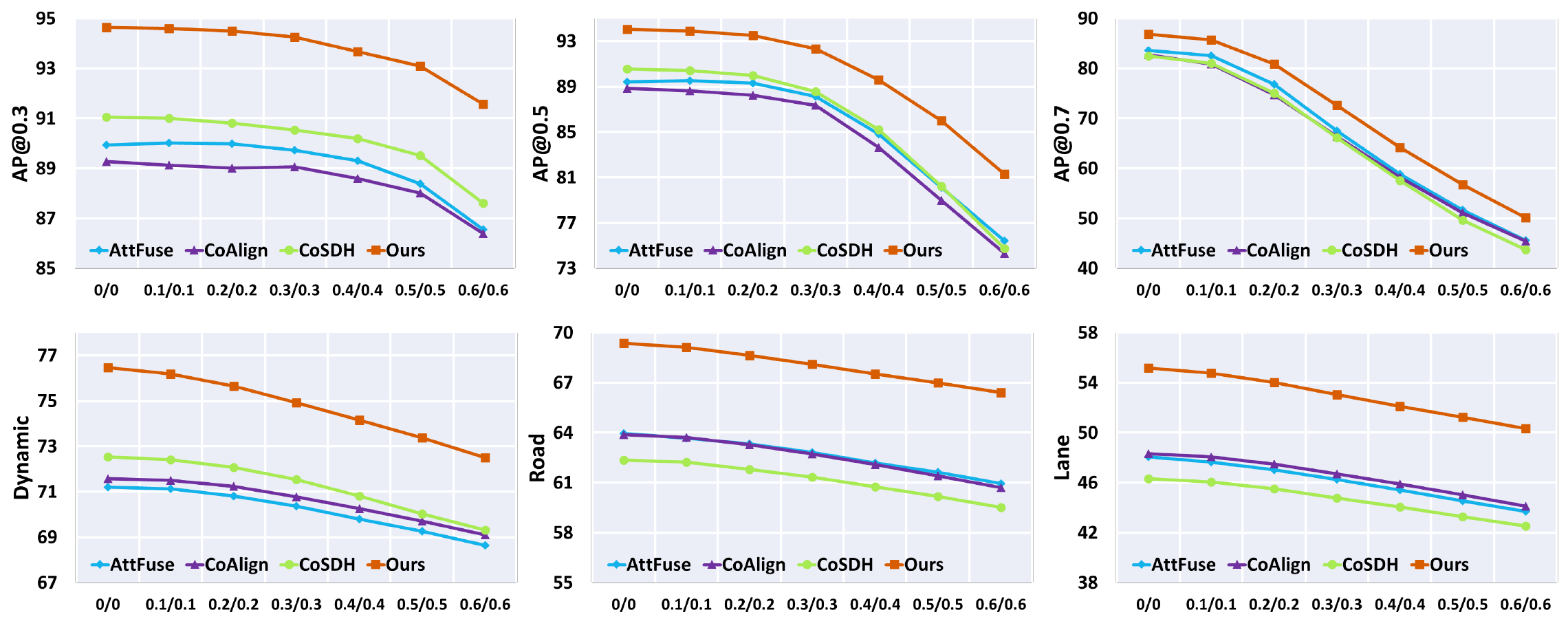}
    \caption{
    Robustness evaluation under increasing Gaussian localization noise on OPV2V.
    The x-axis denotes the injected pose noise level (translation in meters / rotation in degrees),
    ranging from $(0.0\,\text{m},\,0.0^\circ)$ to $(0.6\,\text{m},\,0.6^\circ)$.
    The top row reports detection performance at AP$@$0.3/AP$@$0.5/AP$@$0.7, while the bottom row reports BEV segmentation IoU for Dynamic, Road, and Lane.
    }
    \label{fig:noise}
    \Description{None.}
\end{figure*}

\begin{table*}[!t]
\centering
\caption{Robustness evaluation under varying communication latency on OPV2V.
We sample a random latency from a uniform distribution $\mathbf{U}[0, L]$, gradually increasing the maximum latency $L$ from 200,ms to 500,ms. AP$@$0.5, AP$@$0.7, and Dynamic IoU are reported for each latency setting.
Best results are in \textbf{bold}.}
\label{tab:latency}
\resizebox{\textwidth}{!}{
\begin{tabular}{c | ccc | ccc | ccc | ccc | ccc}
\hline
\multirow{2}{*}{\textbf{Method}} 
& \multicolumn{3}{c|}{\textbf{0\,ms}} 
& \multicolumn{3}{c|}{\textbf{0--200\,ms}} 
& \multicolumn{3}{c|}{\textbf{0--300\,ms}} 
& \multicolumn{3}{c|}{\textbf{0--400\,ms}} 
& \multicolumn{3}{c}{\textbf{0--500\,ms}} \\
\cline{2-16}
& \textbf{AP$@$0.5} & \textbf{AP$@$0.7} & \textbf{Dynamic} 
& \textbf{AP$@$0.5} & \textbf{AP$@$0.7} & \textbf{Dynamic} 
& \textbf{AP$@$0.5} & \textbf{AP$@$0.7} & \textbf{Dynamic} 
& \textbf{AP$@$0.5} & \textbf{AP$@$0.7} & \textbf{Dynamic} 
& \textbf{AP$@$0.5} & \textbf{AP$@$0.7} & \textbf{Dynamic} \\
\hline
AttFuse~\cite{10-attfuse} 
& 89.41 & 83.65 & 71.20 
& 82.34 & 61.03 & 68.12 
& 73.47 & 53.08 & 66.16 
& 67.45 & 49.53 & 64.24 
& 63.56 & 47.29 & 62.33 \\

CoAlign ~\cite{11-coalign} 
& 88.85 & 82.73 & 71.57 
& 81.07 & 58.89 & 68.76 
& 72.02 & 51.89 & 66.58 
& 65.32 & 47.51 & 65.08 
& 61.83 & 45.59 & 63.47 \\

CoSDH~\cite{13-cosdh}    
& 90.54 & 82.48 & 72.54 
& 79.94 & 56.22 & 69.16 
& 69.78 & 48.70 & 66.60 
& 63.19 & 45.24 & 64.66 
& 58.47 & 42.33 & 62.34 \\
\hline
Ours     
& \textbf{94.05} & \textbf{86.88} & \textbf{76.47} 
& \textbf{88.05} & \textbf{63.98} & \textbf{72.95} 
& \textbf{79.47} & \textbf{55.71} & \textbf{70.68} 
& \textbf{72.98} & \textbf{51.63} & \textbf{68.58} 
& \textbf{68.89} & \textbf{50.08} & \textbf{66.50} \\
\hline
\end{tabular}
}
\end{table*}

\textbf{Robustness to Localization Noise.}
Following the noise injection protocol of~\cite{9-where2comm} and~\cite{45-V2X-ViT}, we apply Gaussian noise $\eta$ from $(0.0\,\text{m},\,0.0^\circ)$ to $(0.6\,\text{m},\,0.6^\circ)$ to pose estimates to simulate localization errors. Results are in Fig.~\ref{fig:noise}. As noise grows, all methods degrade on both tasks. However, our CoDS maintains a clear lead over all baselines across all metrics. We believe that this robustness advantage stems from the Collaborative Reliability Map (CoRM) and Semantic Mixture-of-Experts (S-MoE), which perform differentiated processing. This effectively mitigates the destructive effects of noise accumulation on the fused features, enabling CoDS to remain stable even with severe errors.

\textbf{Robustness to Communication Latency.}
To further assess how each method withstands communication delays, we inject random latency sampled from a uniform distribution $\mathbf{U}[0, L]$ and gradually increase the upper bound $L$ from 200\,ms to 500\,ms, with results presented in Table~\ref{tab:latency}. While all methods exhibit varying degrees of performance degradation as latency grows, our CoDS sustains a clear and consistent advantage over every baseline throughout all delay conditions, with a comparatively more gradual performance decline.
Taking the most demanding setting of 500\,ms as an example, CoDS still records AP$@$0.5, AP$@$0.7, and Dynamic IoU of 68.89, 50.08, and 66.50, respectively, retaining a meaningful margin over the best-performing competitor. These findings highlight the stronger perceptual stability of CoDS in realistic communication environments where temporal asynchrony is inevitable.

\textbf{Effect of the Joint Multi-task Training.}
To investigate whether complementary supervision between detection and segmentation can lead to bidirectional performance improvement, we removed the segmentation head and detection head from the Joint Training model, respectively, to obtain a Detection-only model (Det-only) and a Segmentation-only model (Seg-only). The results are shown in Table~\ref{tab:joint}.
Compared with the Det-only, Joint Training achieves consistent improvements across all IoU thresholds. Meanwhile, Joint Training also surpasses the Seg-only across all three semantic categories, with Dynamic, Road, and Lane IoU improving by 5.25, 3.92, and 5.51 points, respectively, validating that joint training effectively drives mutual performance improvements on both tasks.

\begin{table}[!t]
\centering
\caption{Effect of joint multi-task training on OPV2V. 
``Det-only'' and ``Seg-only'' denote models trained with a single task objective, while ``Joint Training'' refers to the full CoDS trained with both tasks simultaneously.
Best in \textbf{bold}.}
\label{tab:joint}
\resizebox{\columnwidth}{!}{%
\begin{tabular}{c | c | ccc}
\hline
\textbf{Module} & \textbf{Test (AP$@$0.3/0.5/0.7)} & \textbf{Dynamic} & \textbf{Road} & \textbf{Lane} \\
\hline
Det-only        & 92.38 / 91.69 / 84.59 & -              & -     & -     \\
Seg-only        & -                      & 71.22          & 65.46 & 49.68 \\
\hline
Joint Training  & \textbf{94.64 / 94.05 / 86.88}  & \textbf{76.47} & \textbf{69.38} & \textbf{55.19} \\
\hline
\end{tabular}
}
\end{table}

\begin{table}[!t]
\centering
\caption{Comparison with single-agent and cooperative multi-task frameworks on OPV2V. BEVerse~\cite{21-beverse} is adapted to the collaborative scenario following the intermediate fusion
strategy of AttFuse~\cite{10-attfuse} for fair comparison. Results for ICA-V2X~\cite{39-ICAV2X} are reproduced from the original paper,
as no official code has been released. Best in \textbf{bold}.} 
\label{tab:multitask}
\resizebox{\columnwidth}{!}{%
\begin{tabular}{c | c | ccc}
\hline
\textbf{Method} & \textbf{AP$@$0.3/0.5/0.7} & \textbf{Dynamic} & \textbf{Road} & \textbf{Lane} \\
\hline
BEVerse~\cite{21-beverse}         & 92.66 / 91.65 / 82.75 & 69.32          & 63.61 & 44.78 \\
CoBEVFusion~\cite{19-cobevfusion} & 93.26 / 92.59 / 83.55 & 73.32          & 59.83 & 45.13 \\
ICA-V2X~\cite{39-ICAV2X}         & - / 92.10 / 84.50     & -              & 64.70 & 53.70 \\
\hline
Ours (CoDS)                       & \textbf{94.64 / 94.05 / 86.88} & \textbf{76.47} & \textbf{69.38} & \textbf{55.19} \\
\hline
\end{tabular}
}
\end{table}

\textbf{Comparison with Multi-task Frameworks.}
Table~\ref{tab:multitask} presents a comprehensive comparison between our CoDS and existing single-agent as well as cooperative multi-task frameworks on OPV2V. Overall, CoDS consistently outperforms all multi-task baselines on both detection and segmentation tasks. In comparison with the single-agent multi-task method BEVerse~\cite{21-beverse}, CoDS achieves notable improvements on both tasks, which we attribute to the explicit modeling of spatial inconsistency in collaborative fusion, allowing CoDS to better exploit the complementary information across multiple vehicles and yield superior perception performance. In comparison with the cooperative multi-task methods CoBEVFusion~\cite{19-cobevfusion} and ICA-V2X~\cite{39-ICAV2X}, our CoDS still maintains a consistent performance lead, with particularly pronounced gains on the Dynamic and Road categories of the segmentation task.  These results further validate the effectiveness of our CoDS in mitigating spatial inconsistency in fusion quality and addressing the limitations of fused features on multi-task representation.

\begin{table}[!t]
\centering
\caption{Comparison of model efficiency. Params., FLOPs, and Mem. denote the parameter count, floating-point operations,
and average GPU memory consumption during inference.
}
\label{tab:efficiency}
\resizebox{0.48\textwidth}{!}{
\begin{tabular}{c | c c c | c | cc}
\hline
\textbf{Method} & \textbf{Params.~$\downarrow$} & \textbf{FLOPs~$\downarrow$} & \textbf{Mem.~$\downarrow$} & \textbf{AP$@$0.5} & \textbf{Dynamic}  & \textbf{Lane} \\
\hline
AttFuse~\cite{10-attfuse}  & 8.1M  & 97.3\,G  & 616.3\,MB  & 89.4 & -                & -     \\
CoBEVT~\cite{12-cobevt}   & 12.2M & 190.8\,G & 573.0\,MB    & -     & 74.0           & 50.8 \\
\hline
NCM      & 20.3M & 381.2\,G & 1336.7\,MB & 89.4 & 74.0   & 50.8 \\
Ours     & 11.0M & 283.9\,G & 673.0\,MB & 94.1 & 76.5  & 55.2 \\
\hline
\end{tabular}
}
\end{table}

\textbf{Efficiency of Joint Multi-task Training.}
Table~\ref{tab:efficiency} evaluates CoDS against single-task models and the Na\"{i}ve Cascade Model~(NCM) across three efficiency dimensions: parameter count, FLOPs, and average GPU memory consumption during inference. NCM is constructed by directly combining the single-task detection model AttFuse~\cite{10-attfuse}
and the single-task segmentation model CoBEVT~\cite{12-cobevt} without any task interaction. Compared with NCM, CoDS reduces the parameter count by 45.8\%, FLOPs by 25.5\%, and inference memory by 49.7\%, while simultaneously achieving superior performance on both detection and segmentation tasks. These results highlight the practical efficiency advantage of CoDS and demonstrate its potential for real-world deployment.

\begin{figure}[!t]
    \centering
    \includegraphics[width=\linewidth]{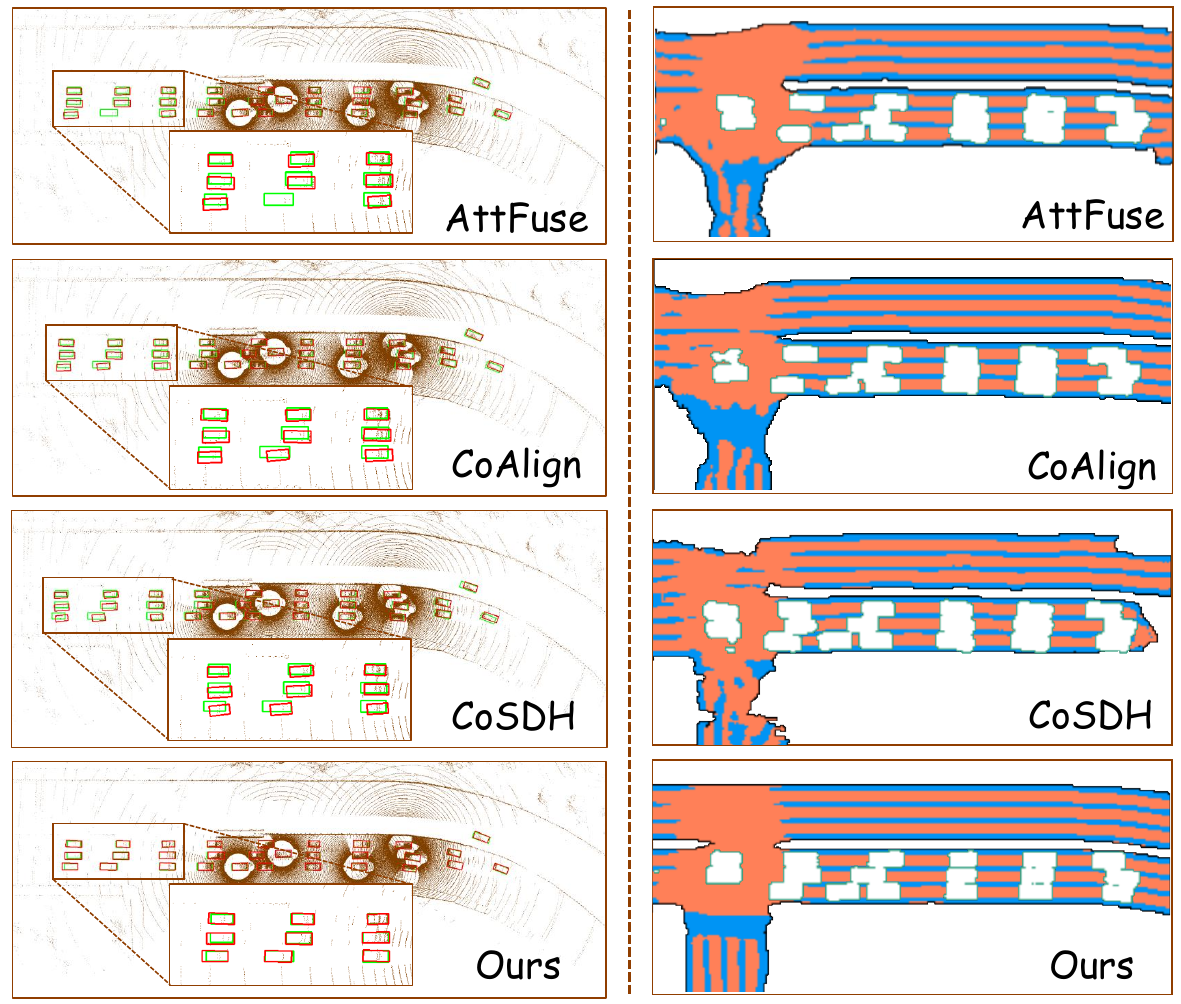}
    \caption{
    Qualitative comparison of detection and BEV segmentation results. Left column: 3D detection results, where \textcolor{green}{green} and \textcolor{red}{red} boxes denote ground truth and predictions, respectively.
    Right column: BEV segmentation results, where \textcolor{OrangeRed}{orangered}, 
    \textcolor{RoyalBlue}{royalblue}, and white regions represent road, lane, and dynamic objects, 
    respectively.
    }
    \label{fig:vis}
    \Description{None.}
\end{figure}

\subsection{Qualitative Analysis}
To show our CoDS robustness under multi-source noise, we inject localization noise of $(0.2\,\text{m},\,0.2^\circ)$ and compare its detection and segmentation visualizations with three baselines (Fig.~\ref{fig:vis}). In detection (left), AttFuse~\cite{10-attfuse}, CoAlign~\cite{11-coalign}, and CoSDH~\cite{13-cosdh} all exhibit obvious missed detections or localization offsets under noise perturbation, whereas CoDS maintains substantially higher spatial alignment between predicted and ground-truth bounding boxes with more complete object recall. In segmentation (right), baselines produce considerable noise fragments and boundary confusion in road and lane regions. Moreover, dynamic objects are often under-segmented or blended with background. In contrast, CoDS yields more structurally coherent segmentation maps with well-defined road boundaries, intact lane markings, and more complete delineation of dynamic objects. These results confirm CoDS achieves more robust perception in both detection and segmentation.


%

\section{Conclusion}
In this paper, we propose \textbf{CoDS}, a robust collaborative multi-task perception framework that jointly optimizes 3D object detection and BEV semantic segmentation under multi-source noise. To handle spatial inconsistency in fusion quality, we introduce CoRM to explicitly quantify the spatial quality distribution of BEV fusion features. Built on CoRM, the S-MoE performs differentiated feature extraction for each task to address the inconsistent feature demands between the two tasks. Moreover, BTCI further mitigates noise-induced feature degradation via bidirectional injection, enabling mutual refinement of task-aware features. Extensive experiments on OPV2V and V2V4Real demonstrate that our CoDS outperforms baselines on both tasks and exhibits strong robustness under localization noise and communication latency perturbations. In summary, our CoDS provides a new paradigm for robust perception and reliable deployment in collaborative systems.

\textbf{Limitation.}
CoDS currently focuses on joint training of 3D object detection and BEV segmentation. A promising direction is integrating additional tasks into the collaborative framework.



\bibliographystyle{ACM-Reference-Format}
\bibliography{main}


\end{document}